\documentclass[10pt,twocolumn,letterpaper]{article}

\usepackage[pagenumbers]{wacv}

\usepackage{multirow}
\usepackage{colortbl}
\usepackage{placeins}

\colorlet{colorFst}{Green!25}
\colorlet{colorSnd}{SpringGreen!45}
\colorlet{colorTrd}{Yellow!30}

\usepackage{acronym}
\newacro{GNSS}{Global Navigation Satellite System}
\newacro{LoD}{Level-of-Detail}
\newacro{PnP}{Perspective-n-Point}
\newacro{UAV}{Unmanned Aerial Vehicle}
\newacro{GPS}{Global Positioning System}
\newacro{IMU}{Inertial Measurement Unit}

\AtBeginDocument{%
  \crefname{figure}{Fig.}{Figs.}%
  \Crefname{figure}{Figure}{Figures}%
  \crefname{table}{Tab.}{Tabs.}%
  \Crefname{table}{Table}{Tables}%
}

\makeatletter
\renewcommand{\NAT@citenum}[3]{%
  \ifNAT@swa
    \mbox{\NAT@@open\if*#2*\else#2\NAT@spacechar\fi
      #1\if*#3*\else\NAT@cmt#3\fi\NAT@@close}%
  \else#1\fi\endgroup}
\makeatother

\definecolor{wacvblue}{rgb}{0.21,0.49,0.74}
\usepackage[pagebackref,breaklinks,colorlinks,allcolors=wacvblue]{hyperref}

\def\confName{WACV}
\def\confYear{2027}

\title{PlanarCityLoc: Breaking Silhouette Ambiguity in LoD\\Aerial Localization via 3D Planes}

\author{%
Oussema Dhaouadi$^{1,2,4,5,6}$\quad Shao Jie Hu Chen$^{1,2}$\quad Jakob Gollreiter$^{1,2}$\\
Samir Jusufi$^{1,3}$\quad Nikolas Rieger$^{1,2,7}$\quad Caroline Wendlandt$^{1,2}$\\
Christian Wender$^{1,8}$\quad Jingfeng Mao$^{1,2,4}$\quad Guangming Wang$^{1,4}$\\
Zuria Bauer$^{5}$\quad Marc Pollefeys$^{5,9}$\quad Daniel Cremers$^{2,6}$\quad Olaf Wysocki$^{1,4}$\\[0.5em]
{\normalsize $^1$TUM.ai\quad $^2$Technical University of Munich\quad $^3$LMU Munich}\\
{\normalsize $^4$University of Cambridge\quad $^5$ETH Z\"urich}\\
{\normalsize $^6$Munich Center for Machine Learning}\\
{\normalsize $^7$Konrad Zuse School of Excellence in Reliable AI (relAI)}\\
{\normalsize $^8$BMW Group\quad $^9$Microsoft}%
}

\makeatletter
\g@addto@macro\@author{\\[0.4em]{\normalsize\url{https://tum-vision.github.io/PlanarCityLoc/}}}
\makeatother

\begin{document}
\maketitle

\begin{abstract}
    Aerial visual localization is crucial for autonomous drone navigation in urban environments where GNSS signals are often degraded. Lightweight textureless Level-of-Detail (LoD) city models provide a geometric map alternative to textured assets. Existing silhouette-based methods rely on image-space alignment, which can leave different camera poses indistinguishable and favor incorrect poses when segmentation errors distort the building outlines. We present PlanarCityLoc, which moves beyond silhouette overlap by leveraging the planar geometry of building facades and roofs. Following geometry-aware initialization, we match image-side planar patches to LoD model planes and use the resulting 3D-to-2D correspondences to refine pose with Perspective-n-Point (PnP). The framework uses off-the-shelf geometry and segmentation models without dataset-specific training or fine-tuning. Experiments on UAVD4L-LoDv2 and Swiss-EPFLv2 achieve state-of-the-art localization recall among the evaluated off-the-shelf LoD-based methods. By grounding image-space alignment in 3D planes, PlanarCityLoc enables more accurate aerial localization with lightweight, textureless city maps.
\end{abstract}
    
\section{Introduction}
\label{sec:intro}

\begin{figure*}[t]
    \centering
    \includegraphics[width=0.85\linewidth]{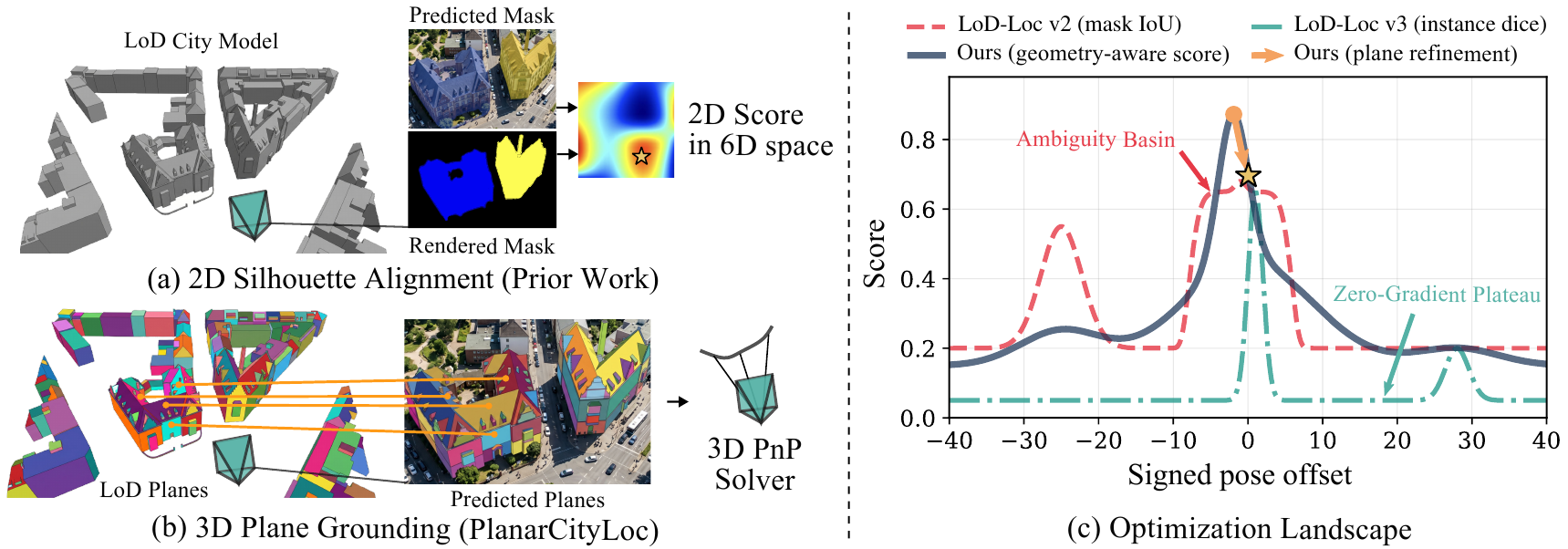}
    \vspace{-0.5\baselineskip}
    \caption{\textbf{PlanarCityLoc overview.}
    (a) Silhouette-based localization compares predicted building masks with renderings of the LoD model. (b) PlanarCityLoc grounds image-side planar patches in 3D through explicit correspondences with LoD planes. (c) Schematic score landscapes illustrate ambiguity in image-space overlap and a score maximum displaced from the correct pose. Plane-based refinement adds geometric constraints to move beyond image-space score maximization. Curves are illustrative, not measured method comparisons.}
    \vspace{-1\baselineskip}
    \label{fig:teaser}
\end{figure*}

Aerial visual localization for \acp{UAV}, defined as estimating the precise 6-Degrees-of-Freedom (6DoF) camera pose from a single image, is essential for operations in \ac{GNSS}-denied urban environments~\cite{villa2020survey, cai2022review}. Camera pose estimates also support geometry-guided annotation of aerial imagery~\cite{gross2026segfly} and semantic mapping for autonomous landing~\cite{gross2026safeland}. While matching images to textured 3D meshes~\cite{schonberger2016structure, wu2024uavd4l} is established, maintaining these assets globally is expensive and privacy-restricted~\cite{nex2024usegeo}. Textureless \ac{LoD} city models provide a compact geometric reference without storing facade appearance~\cite{zhu2024lod, zhu2025lod, peng2026lod, wysocki2024reviewing, groger2012ogc}.

One way to bridge the appearance gap between images and textureless maps is to compare their structural projections. LoD-Loc~\cite{zhu2024lod} follows this approach by predicting building wireframes in the query image and aligning them with edges projected from the LoD model. LoD-Loc v2~\cite{zhu2025lod} instead compares building silhouettes, making alignment applicable to simpler LoD1 geometry (\cref{fig:teaser}(a)). LoD-Loc v3~\cite{peng2026lod} uses instance masks to distinguish buildings, reducing ambiguity and broad score plateaus caused by merged semantic silhouettes in dense scenes.

These silhouette-based objectives still measure agreement in image space. Distinct camera poses can produce similar projections, and disjoint masks yield the same zero intersection-over-union over a range of poses. Segmentation errors introduce a further difficulty. Missing facades, displaced boundaries, or merged building instances change the target silhouette, so an incorrect pose can align better with the predicted mask than the true pose does. Consequently, even a distinct score maximum need not coincide with minimum pose error, as illustrated schematically in \cref{fig:teaser}(c). Improving pose accuracy therefore calls for constraints beyond silhouette overlap alone.

Our key observation is that building facades and roofs provide \textit{explicit 3D planar patches} within \ac{LoD} models. Off-the-shelf vision models~\cite{hu2024metric3d, fu2024geowizard, wang2025moge2, depthanything, kirillov2023segment, ravi2025sam, carion2026sam} provide image-side geometry and segmentation without dataset-specific fine-tuning. Building on these cues, we propose \textit{PlanarCityLoc}. Following geometry-aware coarse initialization, we match image-side planar patches to rendered LoD planes. This \textit{3D planar grounding} uses rendered depth to establish 3D-to-2D correspondences and iteratively refines pose with a robust \ac{PnP} solver~\cite{fischler1981random} (\cref{fig:teaser}(b)). The resulting geometric constraints allow refinement beyond an image-space score peak, addressing the gap between silhouette agreement and accurate camera pose illustrated in \cref{fig:teaser}(c).

In summary, our main contributions are:
\begin{itemize}
    \item We formulate zero-shot \ac{LoD}-based aerial localization through explicit plane correspondences.
    \item We introduce iterative plane association between image patches and textureless map geometry for robust refinement of the camera pose.
    \item We evaluate on UAVD4L-LoDv2 and Swiss-EPFLv2, achieving state-of-the-art localization recall among the evaluated off-the-shelf LoD-based methods across all four splits, and analyze the contributions of geometric scoring and plane-based refinement through ablations.
\end{itemize}

\section{Related Work}
\label{sec:related}

\noindent\textbf{Visual localization.}
Visual localization estimates the 6DoF camera pose relative to a known map. Traditional approaches match local features to a 3D point cloud constructed via Structure-from-Motion~\cite{schonberger2016structure, pan2024global}, pairing feature extractors~\cite{detone2018superpoint} with robust matchers~\cite{sarlin2020superglue, sun2021loftr, edstedt2024roma, edstedt2025romav2}. Hierarchical pipelines~\cite{sarlin2019coarse} first retrieve relevant images before local feature matching. MeshLoc~\cite{panek2022meshloc} localizes directly on 3D meshes, while UAVLoc~\cite{wu2024uavd4l} uses posed reference images rendered from textured meshes for aerial localization. Render-and-compare approaches~\cite{yan2023render} offer another route to aerial pose estimation. MC-Loc~\cite{trivigno2024unreasonable} combines pretrained image features, a renderable scene representation, and particle filtering for pose refinement without task-specific training. Maintaining detailed appearance-bearing maps at city scale remains costly. AnyVisLoc~\cite{ye2026anyvisloc}, OrthoLoC~\cite{dhaouadi2025ortholoc}, and OrthoTrack~\cite{dhaouadi2026orthotrack} instead explore localization using overhead imagery and 2.5D surface maps. Our work focuses on textureless \ac{LoD} geometry, which provides a compact reference without storing facade appearance.

\noindent\textbf{LoD-based localization.}
LoD-Loc~\cite{zhu2024lod} introduced the use of LoD city models for aerial localization through neural wireframe alignment. It predicts building edges in the query image, scores pose hypotheses using projected map wireframes, and refines the estimate through differentiable alignment. LoD-Loc v2~\cite{zhu2025lod} shifts from wireframes to semantic building silhouettes, supporting simpler LoD1 models through coarse pose search and particle-filter refinement. LoD-Loc v3~\cite{peng2026lod} further separates buildings using instance segmentation trained on synthetic aerial imagery. Instance-level alignment reduces the ambiguity of merged semantic masks in dense scenes, but remains sensitive to missing instances, incorrect boundaries, and merged or split predictions. Such errors can displace the overlap maximum from the true pose, even when the score has a distinct peak (\cref{fig:teaser}(c)). Concurrent work, MatchCityLoc~\cite{dhaouadi2027matchcityloc}, explores an alternative through matching predicted and rendered surface normals. PlanarCityLoc complements silhouette alignment with explicit 3D plane correspondences. This grounding enables refinement beyond the image-space score peak, leveraging pretrained geometry and segmentation models without dataset-specific training or fine-tuning.

\noindent\textbf{Monocular foundation models.}
Extracting structured planar patches from a single \ac{UAV} image requires precise surface geometry and semantic segmentation. Recently, monocular geometry estimation has experienced a paradigm shift due to large-scale foundation models~\cite{depthanything, hu2024metric3d, fu2024geowizard, ke2025marigold}. Models such as MoGe-2~\cite{wang2025moge2} can predict metric-scale 3D point maps and detailed surface normals from a single image without any retraining. Similarly, segmentation foundation models like SAM~3~\cite{carion2026sam} provide robust zero-shot building masks. The OccuFly benchmark~\cite{gross2026occufly} further highlights the challenges of applying pretrained metric-depth models to aerial imagery across different flight altitudes. We leverage these models to extract geometric and semantic cues from aerial images.

\noindent\textbf{Plane-based localization.}
Planes provide structural constraints beyond individual keypoints and have been explored for augmented reality~\cite{servant2008visual}. In indoor localization, PlanaReLoc~\cite{ye2026planareloc} matches query-image and 3D-map planes through learned embeddings. PALMS~\cite{cheng2024palms} uses observed wall geometry and floorplans to initialize indoor particle-filter localization. Related primitive-based methods include Z-FLoc~\cite{umemura2026z}, which matches lines and circles extracted from monocular reconstructions to floorplans, and outdoor approaches that align semantic or geometric primitives across views~\cite{li2021bsp, peterson2026meridian}. These works establish the value of structural primitives, while their use for aerial localization against simplified city models remains less explored. We match facade and roof planes in \ac{LoD} models to patches predicted from a single UAV image.

\section{Method}
\label{sec:method}

\noindent\textbf{Problem formulation.}
Given an \ac{LoD} city model $\mathcal{M}$, a query image ${I_q \in \mathbb{R}^{H \times W \times 3}}$ with height $H$ and width $W$, known intrinsics $\mathbf{K}$, and a coarse GPS/IMU pose prior
$\boldsymbol{\xi}_p$, our goal is to estimate the accurate 6-DoF camera pose
${\boldsymbol{\xi}^* = (x, y, z, \theta, \rho, r)}$, where $(x, y, z)$ is the world-space camera center and $(\theta,\rho,r)$ denotes yaw, pitch, and roll.

\begin{figure*}[t]
  \centering
  \includegraphics[width=\linewidth]{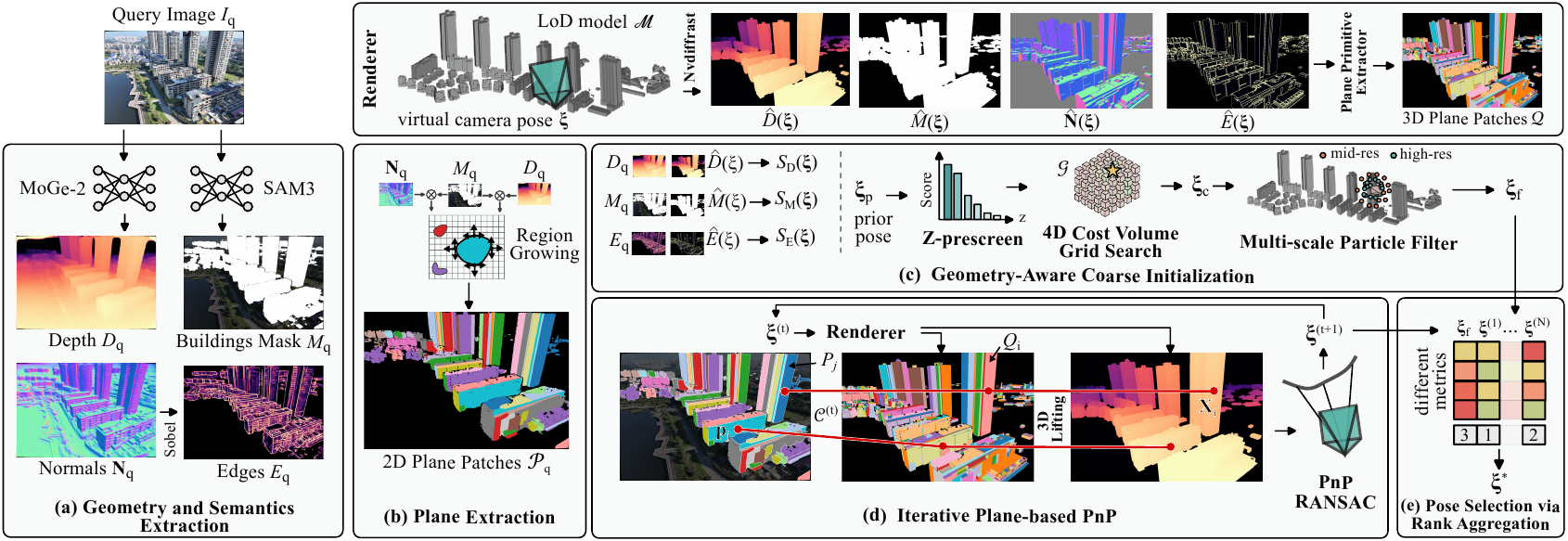}
   \vspace{-1\baselineskip}
  \caption{\textbf{Overview of PlanarCityLoc.}
    (a) We extract building masks, surface normals, depth, and structural edges from query image $I_q$ with foundation models, then (b) form 2D planar patches.
    (c) A coarse pose $\boldsymbol{\xi}_f$ is obtained through altitude initialization, cost-volume search, and multi-scale particle filtering.
    (d) We iteratively match visible LoD planes to image patches and refine the pose using PnP-RANSAC on the resulting 3D-to-2D correspondences.
    (e) Equal-weight rank aggregation of complementary metrics selects the final pose.}
    \vspace{-1\baselineskip}
  \label{fig:pipeline}
\end{figure*}

\noindent\textbf{Overview.}
We show the pipeline in \cref{fig:pipeline}. We first extract geometric cues from the query image and group building surfaces into planar patches. Geometry-aware coarse initialization then brings the image and map into approximate alignment. Building on this estimate, our 3D planar grounding stage matches image patches to visible LoD planes and iteratively refines the pose through robust \ac{PnP}. Finally, rank aggregation combines complementary alignment measures to select the output pose.

\subsection{Geometry and Semantics Extraction}
\label{sec:extraction}

We extract query-side geometry and semantics using off-the-shelf foundation models. MoGe-2~\cite{wang2025moge2} provides a surface-normal map ${\mathbf{N}_q \in \mathbb{R}^{H \times W \times 3}}$ and a depth map ${D_q \in \mathbb{R}^{H \times W}_{>0}}$, while SAM~3~\cite{carion2026sam} provides a binary building mask ${M_q \in \{0,1\}^{H \times W}}$. We derive a structural edge map $E_q$ by applying a Sobel filter to $\mathbf{N}_q$ within a dilated $M_q$. These edges capture changes in surface orientation within and around buildings. We denote the unit normal at pixel $u$ as ${\mathbf{n}_q(u) = \mathbf{N}_q(u)/\|\mathbf{N}_q(u)\|}$.

\subsection{Plane Extraction}
\label{sec:plane_extract}

The predicted geometry allows us to describe buildings through their individual facade and roof surfaces, rather than a single silhouette. These surfaces have counterparts in the planar geometry of LoD city models. We extract patches from the query image and rendered map views to establish correspondences for pose refinement.

\noindent\textbf{Image-side patchification.}
Starting from the building mask $M_q$, we group pixels into planar patches $\mathcal{P}_q=\{P_j\}$ using 4-connected region growing. We merge neighboring pixels $u$ and $v$ when their normals agree,
\begin{equation}
  \mathbf{n}_q(u) \cdot \mathbf{n}_q(v) \ge \tau_n,
\label{eq:plane_normal}
\end{equation}
where $\tau_n$ is a normal similarity threshold. Furthermore, we require their relative depths to be consistent,
\begin{equation}
  \frac{|D_q(u) - D_q(v)|}{\max\!\left(D_q(u),\, D_q(v)\right)} < \tau_d,
\label{eq:plane_edge}
\end{equation}
where $\tau_d$ is a relative depth threshold.
Depth continuity helps separate adjacent surfaces with similar normals but different ranges. Each patch $P_j = \big(U_j,\, \mathbf{p}_j,\, \bar{\mathbf{n}}_j\big)$ comprises its pixel support $U_j$, a robust 2D center $\mathbf{p}_j$ at the median pixel location, and a normalized mean normal $\bar{\mathbf{n}}_j$.

\noindent\textbf{Model-side patchification.}
Given a pose hypothesis $\boldsymbol{\xi}$, we render the mask $\hat{M}(\boldsymbol{\xi})$, depth $\hat{D}(\boldsymbol{\xi})$, and normal map $\hat{\mathbf{N}}(\boldsymbol{\xi})$, and derive rendered edges $\hat{E}(\boldsymbol{\xi})$ for scoring. We apply the same patch-extraction rule to obtain visible LoD patches $\mathcal{Q}(\boldsymbol{\xi})=\{Q_i\}$. Each patch $Q_i=\big(V_i(\boldsymbol{\xi}),\mathbf{X}_i,\hat{\mathbf{N}}_i\big)$ contains its pixel support, a representative 3D center, and its camera-frame unit normal. We obtain $\mathbf{X}_i$ by back-projecting the center of the visible patch using rendered depth and transforming it to world coordinates. Matching 3D centers to image locations supplies the correspondences for \ac{PnP}.

\subsection{Geometry-Aware Coarse Initialization}
\label{sec:coarse}

Reliable plane matching first requires approximate alignment between the query image and map. We obtain this initialization by rendering pose hypotheses around the sensor prior $\boldsymbol{\xi}_p$ and comparing them with the query predictions. Following the image-space search paradigm of LoD-Loc v2 and LoD-Loc v3~\cite{zhu2025lod,peng2026lod}, we use cost-volume (CV) search followed by particle-filter (PF) refinement, augmented with edge and depth cues. Both stages refine translation and yaw while keeping pitch and roll fixed at their prior estimates, as in LoD-Loc v2~\cite{zhu2025lod}. The subsequent plane-based PnP stage refines all six pose degrees of freedom.

\noindent\textbf{Cost-volume search.}
We first sample altitude hypotheses at the prior horizontal location to initialize height. We then search a translation-and-yaw grid $\mathcal{G}$ in two successive stages, narrowing the grid around the best hypothesis. The cost-volume alignment score $S_{\mathrm{CV}}$ combines building-mask intersection-over-union $S_I=|M_q\cap\hat M|/|M_q\cup\hat M|$, edge agreement $S_E$, and depth consistency $S_D$. The edge term rewards rendered building boundaries that lie close to query normal-map edges, weighted by their coverage. The depth term combines similarity in depth variation with residual error after compensating for scale and offset differences. We express this cue fusion in the general weighted-sum form
\begin{equation}
  S_{\mathrm{CV}}(\boldsymbol{\xi}) =
    w_I^c(\boldsymbol{\xi})\,S_I(\boldsymbol{\xi})
    + w_E^c(\boldsymbol{\xi})\,S_E(\boldsymbol{\xi})
    + w_D^c(\boldsymbol{\xi})\,S_D(\boldsymbol{\xi}).
\label{eq:cv_score}
\end{equation}
Here, the cue-dependent weights $w_I^c$, $w_E^c$, and $w_D^c$ balance mask, edge, and depth agreement, respectively. On each grid, selection applies a weak proximity prior and a mask-overlap floor to favor plausible hypotheses. We denote the resulting coarse pose by $\boldsymbol{\xi}_c$.

\noindent\textbf{Particle-filter refinement.}
Starting from $\boldsymbol{\xi}_c$, a multi-scale particle filter~\cite{karkus2018particle,trivigno2024unreasonable} samples local pose perturbations and favors hypotheses with stronger rendered-to-query agreement. The perturbation scale decreases over successive steps to refine the pose. This stage uses the additive alignment score
\begin{equation}
  S_{\mathrm{PF}}(\boldsymbol{\xi}) =
    w_m \, S_M(\boldsymbol{\xi})
    + w_e \, S_E(\boldsymbol{\xi})
    + w_d \, S_D(\boldsymbol{\xi}).
\label{eq:render_score}
\end{equation}
Here, $w_m$, $w_e$, and $w_d$ balance the three cues. Unlike the cost volume's IoU term, $S_M$ measures the fraction of the predicted building mask explained by the rendering:
\begin{equation}
  S_M(\boldsymbol{\xi}) = \frac{|M_q \cap \hat{M}(\boldsymbol{\xi})|}{|M_q|}.
\label{eq:mask_score}
\end{equation}
Mask recall tolerates building regions present in the rendered map but missing from the predicted mask, for example because of occlusion or incomplete segmentation. Unlike IoU, it does not penalize this extra rendered support. Edge and depth agreement provide complementary constraints, with a weak proximity prior discouraging drift during local search. The best pose retained after the final particle-filter stage, $\boldsymbol{\xi}_f$, initializes plane-based refinement.

\subsection{Iterative Plane-based PnP}
\label{sec:pnp}

The coarse estimate $\boldsymbol{\xi}_f$ aligns the rendered map with the image, but image-space agreement can still leave residual pose ambiguity. We refine this estimate by grounding image patches in the map's 3D geometry. Each matched LoD patch provides a world-space point $\mathbf{X}_i$ and a corresponding image location $\mathbf{u}_i$. Together, these 3D-to-2D correspondences constrain the camera pose through \ac{PnP}, allowing refinement beyond the image-space score peak.

\noindent\textbf{Iterative PnP.}
Starting from $\boldsymbol{\xi}^{(0)}=\boldsymbol{\xi}_f$, we re-render the LoD planes at each outer iteration $t$ and construct a set of 3D-to-2D correspondences $\mathcal{C}^{(t)}=\{(\mathbf{X}_i,\mathbf{u}_i)\}$. We update the pose as
\begin{equation}
  \boldsymbol{\xi}^{(t+1)} = \operatorname{PnP}\!\left(\mathbf{K},\mathcal{C}^{(t)}\right)
\label{eq:pnp}
\end{equation}
using PnP-RANSAC~\cite{fischler1981random} followed by nonlinear refinement on the inliers. Patch centers can be imprecise because of segmentation errors or partial visibility. Given sufficient consistent correspondences, RANSAC rejects outlying matches so that every center need not be accurate. We repeat matching and pose estimation for up to $N$ outer iterations, re-rendering each updated pose to renew correspondences and gather pose candidates.

\noindent\textbf{Correspondence building.}
Image-side extraction can split a physical facade into several patches, while its rendered counterpart remains a single patch. To accommodate this fragmentation and then refine the associations, we apply two sequential matchers within each iteration.

\emph{Overlap voting} associates each rendered patch with the query fragments that overlap it. Let $n_{ij}$ count the overlapping pixels between rendered patch $Q_i$ and query patch $P_j$. We require a dominant overlap vote and consistent rendered and mean fragment normals:
\begin{equation}
  j = \arg\max_{j'} n_{ij'},
  \quad
  \frac{n_{ij}}{\sum_{j'} n_{ij'}} \ge \tau_v,
  \quad
  \hat{\mathbf{N}}_i^\top \bar{\mathbf{n}}_{\mathrm{merged},i} \ge \tau_c,
\label{eq:overlap_vote}
\end{equation}
where $\tau_v$ is the minimum vote fraction, $\tau_c$ is the normal similarity threshold, and $\bar{\mathbf{n}}_{\mathrm{merged},i}$ is the normalized mean normal of the supporting fragments. The target $\mathbf{u}_i$ is the centroid of their combined overlapping pixels. This merges fragmented image evidence into a single correspondence for each rendered patch, from which we estimate the pose using iterative PnP within RANSAC.

\emph{Centroid consistency} then seeks one-to-one associations between projected map-patch centers and query-patch centers. Using the perspective-projection operator $\pi$, we project $\mathbf{X}_i$ to $\tilde{\mathbf{p}}_i=\pi(\mathbf{K},\boldsymbol{\xi},\mathbf{X}_i)$ and form mutual nearest-neighbor pairs with $\mathbf{p}_j$, which become the targets $\mathbf{u}_i$. Distance and nearest-neighbor ratio tests reject distant or ambiguous pairs. We estimate the pose from these correspondences using SQPnP~\cite{terzakis2020sqpnp} within RANSAC.

\subsection{Final Pose Selection via Rank Aggregation}
\label{sec:selection}

The particle-filter pose and the successive PnP estimates form a candidate set $\mathcal{V}$. A pose with high silhouette overlap need not have the most accurate geometric alignment, while prediction noise can lower the overlap of a well-aligned pose. We therefore combine eight complementary criteria: composite alignment score, mask recall, mask IoU, edge alignment, depth correlation, normal error, reprojection error, and matched-plane centroid distance. The composite alignment score follows the additive form in \cref{eq:render_score}, but uses rendered normal-map edges to also capture internal facade and roof structure. Final selection balances all eight criteria instead of maximizing the composite score alone.

Depth correlation measures normalized cross-correlation between predicted and rendered depth over valid building-mask overlap. Normal error averages angular differences over common building support. Reprojection error measures the mean image-space residual of the fitted inlier correspondences, while centroid distance averages the separation of matched image and rendered patch centers.

Rank aggregation balances these criteria without directly combining their different numerical scales. Let $r_k(\boldsymbol{\xi})$ be the candidate's rank under criterion $k$, with higher ranks assigned to stronger agreement or smaller errors. We select
\begin{equation}
\boldsymbol{\xi}^*=\arg\max_{\boldsymbol{\xi}\in\mathcal{V}}
\frac{1}{8}\sum_{k=1}^{8} r_k(\boldsymbol{\xi}).
\label{eq:rank_selection}
\end{equation}
Each criterion contributes equally, without learning aggregation weights.

\section{Experiments}
\label{sec:Experiments}

We evaluate PlanarCityLoc against published references and adapted LoD baselines. Additional results and evaluation details are provided in the supplementary.

\begin{table*}[t]
\centering
\setlength{\tabcolsep}{3pt}
\renewcommand{\arraystretch}{1.08}
\resizebox{\linewidth}{!}{%
\begin{tabular}{l l c ccc c ccc c ccc c ccc c}
\toprule
 & & & \multicolumn{8}{c}{\textbf{UAVD4L-LoDv2 query splits}} & \multicolumn{8}{c}{\textbf{Swiss-EPFLv2 query splits}} \\
\cmidrule(lr){4-11} \cmidrule(lr){12-19}
 & & & \multicolumn{4}{c}{\textit{in-Traj.}} & \multicolumn{4}{c}{\textit{out-of-Traj.}} & \multicolumn{4}{c}{\textit{in-Place}} & \multicolumn{4}{c}{\textit{out-of-Place}} \\
\cmidrule(lr){4-7} \cmidrule(lr){8-11} \cmidrule(lr){12-15} \cmidrule(lr){16-19}
 \textbf{Method} & \textbf{Vision Model} & \textbf{Setup} & 2m-2$^\circ$ & 3m-3$^\circ$ & 5m-5$^\circ$ & T.e./R.e. & 2m-2$^\circ$ & 3m-3$^\circ$ & 5m-5$^\circ$ & T.e./R.e. & 2m-2$^\circ$ & 3m-3$^\circ$ & 5m-5$^\circ$ & T.e./R.e. & 2m-2$^\circ$ & 3m-3$^\circ$ & 5m-5$^\circ$ & T.e./R.e. \\
\midrule
Released priors & -- & -- & 0.0 & 0.0 & 4.3 & 6.48/1.63 & 0.0 & 0.0 & 0.4 & 11.12/1.33 & 0.0 & 0.0 & 0.6 & 17.59/3.86 & 0.0 & 0.0 & 1.1 & 17.88/3.94 \\
\midrule
\multicolumn{19}{l}{\textbf{Textured Mesh-based Models}} \\
UAVLoc~\cite{wu2024uavd4l} & LoFTR & O & \underline{84.98} & \underline{88.09} & \underline{88.90} & \underline{0.81}/\underline{0.29} & \underline{91.56} & \underline{92.02} & \underline{92.11} & \underline{0.79}/\textbf{0.20} & \underline{26.40} & \underline{46.21} & \underline{62.22} & \underline{3.06}/\underline{1.86} & \underline{68.87} & \underline{81.00} & \underline{84.96} & \underline{1.12}/\underline{1.08} \\
UAVLoc~\cite{wu2024uavd4l} & RoMa & O & \textbf{93.27} & \textbf{93.70} & \textbf{93.77} & \textbf{0.78}/\textbf{0.25} & \textbf{95.03} & \textbf{95.53} & \textbf{95.53} & \textbf{0.73}/\underline{0.22} & \textbf{45.95} & \textbf{66.77} & \textbf{80.73} & \textbf{2.08}/\textbf{1.29} & \textbf{89.18} & \textbf{89.68} & \textbf{98.84} & \textbf{0.77}/\textbf{1.04} \\
\midrule
\multicolumn{19}{l}{\textbf{LoD-based Models}} \\
\multicolumn{19}{l}{\hspace{1em}\textit{Requires Training}} \\
LoD-Loc~\cite{zhu2024lod} & Custom U-Net & I & 49.56 & 71.82 & 89.09 & 3.32/1.48 & 54.20 & 75.05 & 89.51 & 3.33/1.18 & 36.79 & 50.56 & 69.77 & 2.87/1.78 & 14.24 & 31.39 & 59.89 & 8.73/2.78 \\
LoD-Loc v2~\cite{zhu2025lod} & SAM2 + U-Net & I & \underline{93.70} & \underline{98.40} & \underline{99.50} & \underline{0.72}/\underline{0.15} & \textbf{97.90} & \textbf{99.80} & \textbf{100.00} & \underline{0.71}/\underline{0.14} & \underline{54.20} & \underline{74.60} & \underline{92.00} & \underline{1.83}/\underline{0.85} & \underline{31.40} & \underline{58.53} & \underline{86.30} & \underline{2.64}/\textbf{0.73} \\
LoD-Loc v3~\cite{peng2026lod} & SAM-Prompter & C* & \textbf{97.60} & \textbf{98.90} & \textbf{99.70} & \textbf{0.49}/\textbf{0.13} & \underline{97.40} & \underline{99.00} & \underline{99.40} & \textbf{0.60}/\textbf{0.12} & \textbf{58.60} & \textbf{79.90} & \textbf{95.40} & \textbf{1.61}/\textbf{0.77} & \textbf{36.90} & \textbf{64.10} & \textbf{88.90} & \textbf{2.43}/\underline{0.83} \\
\midrule
\multicolumn{19}{l}{\hspace{1em}\textit{Oracle (GT masks)}} \\
LoD-Loc v2$\dagger$~\cite{zhu2025lod} & -- & GT & \textbf{57.2} & \textbf{78.8} & \textbf{92.8} & \textbf{1.85}/\textbf{1.23} & \textbf{64.9} & \textbf{94.5} & \textbf{98.3} & \textbf{1.76}/\textbf{0.94} & \textbf{74.9} & \textbf{90.6} & \textbf{93.5} & \textbf{1.36}/\textbf{0.85} & \textbf{43.3} & \textbf{75.2} & \textbf{93.9} & \textbf{2.15}/\textbf{0.78} \\
LoD-Loc v3$\dagger$~\cite{peng2026lod} & -- & GT & \underline{54.5} & \underline{78.7} & \underline{90.6} & \underline{1.87}/\textbf{1.23} & \underline{41.3} & \underline{83.8} & \underline{96.4} & \underline{2.15}/\underline{0.95} & \underline{56.3} & \underline{72.2} & \underline{78.9} & \underline{1.79}/\underline{0.91} & \underline{40.1} & \underline{73.6} & \textbf{93.9} & \underline{2.27}/\underline{0.79} \\
\midrule
\multicolumn{19}{l}{\hspace{1em}\textit{Off-the-shelf}} \\
MC-Loc~\cite{trivigno2024unreasonable} & DINOv2 & O & 1.20 & 4.10 & 17.40 & 8.29/2.58 & 2.40 & 7.40 & 26.10 & 7.02/2.29 & 0.90 & 4.40 & 17.50 & 8.18/2.54 & 2.90 & 9.00 & 30.20 & 6.23/1.97 \\
MC-Loc~\cite{trivigno2024unreasonable} & RoMa & O & 0.10 & 0.60 & 3.30 & 10.60/8.60 & 0.20 & 0.90 & 3.30 & 16.90/3.88 & 0.20 & 1.20 & 4.80 & 9.80/2.65 & 0.70 & 2.10 & 11.50 & 10.30/3.97 \\
LoD-Loc v2$\dagger$~\cite{zhu2025lod} & SAM3~\cite{carion2026sam} & O & 35.7 & \underline{67.8} & 81.5 & \underline{2.33}/\underline{1.27} & \underline{64.3} & \underline{93.3} & 98.2 & \underline{1.77}/\underline{0.95} & 9.4 & 19.9 & 39.7 & 6.51/1.83 & \underline{17.4} & \underline{33.0} & 66.2 & \underline{3.90}/1.33 \\
\multicolumn{19}{l}{\textit{PlanarCityLoc (Ours)}} \\
\quad Cost Vol. Selection & MoGe-2 + SAM3~\cite{carion2026sam} & O & 27.6 & 59.9 & 79.4 & 2.62/1.49 & 34.9 & 74.1 & 95.3 & 2.35/1.22 & 15.4 & 30.2 & 48.7 & 5.04/\underline{1.72} & 11.3 & 23.2 & 57.5 & 4.42/1.41 \\
\quad + Particle Filter & MoGe-2 + SAM3~\cite{carion2026sam} & O & \underline{36.0} & 67.7 & \underline{82.8} & 2.36/1.30 & 60.8 & 90.6 & \underline{98.5} & 1.81/0.98 & \textbf{24.9} & \textbf{41.6} & \textbf{61.8} & \textbf{3.54}/\textbf{1.54} & 14.5 & 31.1 & \underline{70.2} & 3.99/\underline{1.23} \\
\quad \textbf{+ Plane-based PnP} & MoGe-2 + SAM3~\cite{carion2026sam} & O & \textbf{75.4} & \textbf{90.8} & \textbf{97.2} & \textbf{1.28}/\textbf{0.68} & \textbf{80.6} & \textbf{95.1} & \textbf{99.0} & \textbf{1.26}/\textbf{0.69} & \underline{20.6} & \underline{39.0} & \underline{59.7} & \underline{3.70}/1.90 & \textbf{48.5} & \textbf{72.3} & \textbf{89.4} & \textbf{1.98}/\textbf{1.21} \\
\bottomrule
\end{tabular}%
}
\vspace{-0.5\baselineskip}
\caption{\textbf{Quantitative comparison on UAVD4L-LoDv2 and Swiss-EPFLv2.} Setup: \textbf{O}ff-the-shelf, \textbf{I}n-domain training, \textbf{C}ross-dataset training, or \textbf{GT} oracle masks. \textbf{C}$^*$ uses the InsLoD-Loc training set~\cite{peng2026lod}. Recall is in \%. T.e./R.e. are median translation (m) and rotation ($^\circ$) errors. For evaluated runs, recall requires both pose thresholds. Best/second-best distinct values per column within each setup group are bold/underlined. $\dagger$: adapted re-runs (see \cref{sec:setting_baselines}).}
\vspace{-1\baselineskip}
\label{tab:main_results}
\end{table*}

\subsection{Experiment Settings and Baselines}
\label{sec:setting_baselines}

\noindent \textbf{Datasets.}
We evaluate on the UAVD4L-LoDv2 and Swiss-EPFLv2 query splits~\cite{zhu2025lod,peng2026lod}, inherited from UAVD4L-LoD and Swiss-EPFL~\cite{zhu2024lod}, respectively. These dataset extensions introduce LoD1 maps while retaining the original queries. Each query benchmark has two splits: in-Traj. and out-of-Traj. for UAVD4L-LoDv2, and in-Place and out-of-Place for Swiss-EPFLv2. Swiss-EPFL imagery and poses originate from CrossLoc~\cite{yan2022crossloc}. Our main experiments use LoD3 for UAVD4L-LoDv2 and LoD2 for Swiss-EPFLv2 unless stated otherwise. The released UAVD4L-LoDv2 priors are sensor-derived, whereas Swiss-EPFLv2 priors were synthesized from ground truth by the benchmark authors~\cite{zhu2024lod}.

\noindent \textbf{Baselines.}
\Cref{tab:main_results} includes textured-map methods and LoD-based methods. We report published references alongside adapted LoD-Loc evaluations ($\dagger$).

The adapted evaluations assess localization with oracle building masks and off-the-shelf segmentation, separately from published trained configurations. We use rasterized map renderings with ground-truth-rendered masks for LoD-Loc v2 and LoD-Loc v3, and additionally evaluate LoD-Loc v2 with SAM~3~\cite{carion2026sam} predictions. These variants contrast ideal masks with general-purpose predictions for localization.

\noindent \textbf{Evaluation metrics.}
Following prior work~\cite{zhu2025lod,peng2026lod}, we report median translation and rotation errors, and recall at 2m--2$^\circ$, 3m--3$^\circ$, and 5m--5$^\circ$. A query is recalled when both errors are strictly below the corresponding thresholds.

\subsection{Implementation Details}
\label{sec:Implementation_Details}
The main method uses SAM~3~\cite{carion2026sam} for building masks and MoGe-2 ViT-S~\cite{wang2025moge2} for depth and normals. Plane extraction uses normal threshold $\tau_n=0.995$ and relative depth thresholds $\tau_d=0.05$ for image patches and $0.02$ for rendered patches. We discard patches smaller than 50 pixels.

Coarse localization uses two translation-and-yaw grids followed by medium and fine particle-filter stages, with up to 15 and 6 iterations. Plane refinement uses up to $N=5$ outer iterations. Standard correspondence filtering requires normal consistency $\tau_c=0.7$ and at least six matches, with a $20$\,px RANSAC threshold. Strongly fragmented image patches trigger stricter matching, while insufficient building coverage skips PnP. Search configurations and matching thresholds were established empirically during development and fixed for the reported evaluations.

\subsection{Evaluation Results}
\label{sec:Evaluation_Results}

\noindent \textbf{Quantitative results.}
We report the results in \cref{tab:main_results}. PlanarCityLoc outperforms the evaluated prior off-the-shelf LoD-based methods in recall across all four splits and all three thresholds. Explicit plane correspondences substantially improve the particle-filter estimates on both UAVD4L-LoDv2 splits and Swiss-EPFLv2 out-of-Place. Textured-map UAVLoc remains stronger, benefiting from richer appearance cues unavailable in textureless geometry.

Swiss-EPFLv2 in-Place is more challenging for plane refinement, with 2m--2$^\circ$ recall changing from 24.9\% to 20.6\%. This split has fewer visible map patches and a larger imbalance between predicted and rendered patch counts than out-of-Place, consistent with less reliable plane associations. Refinement also produces fewer successful PnP proposals, and candidate selection sometimes favors an inaccurate update. These observations highlight the importance of sufficiently rich, consistent planar evidence.

\begin{figure*}[t]
\centering
\includegraphics[width=0.96\linewidth]{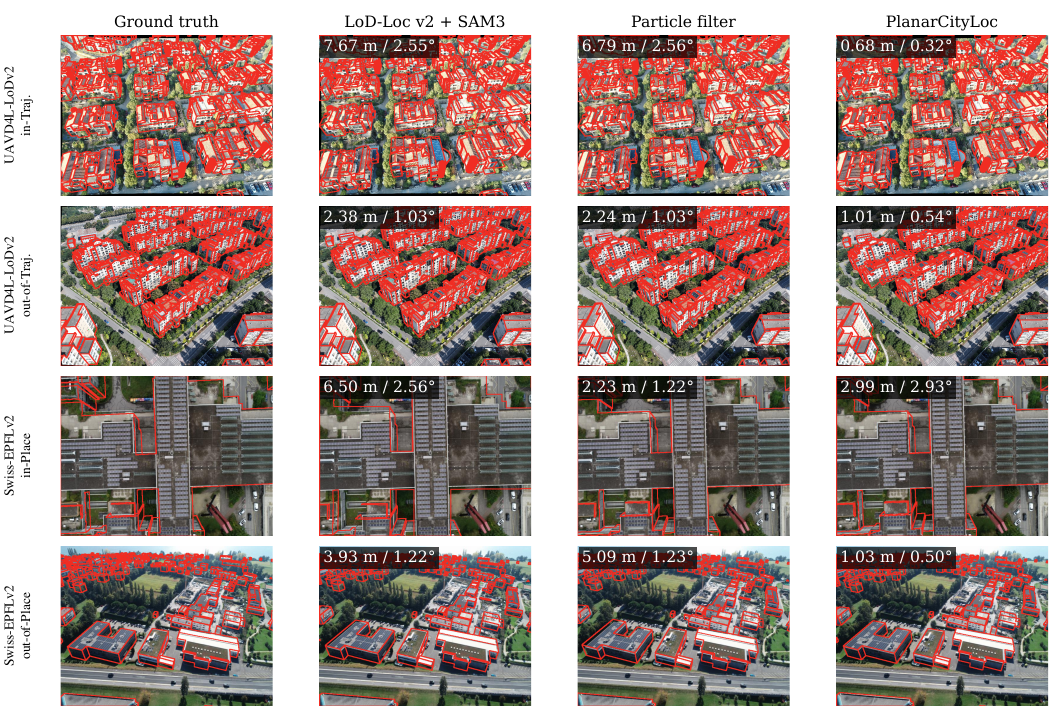}
\caption{\textbf{Qualitative comparison on UAVD4L-LoDv2 and Swiss-EPFLv2.} Red overlays show LoD edges at each estimated pose. Labels give translation (m) / rotation ($^\circ$) errors. Similar visual alignments can correspond to different pose errors.}
\label{fig:iterations}
\end{figure*}

\begin{figure*}[t]
\centering
\includegraphics[width=0.98\linewidth]{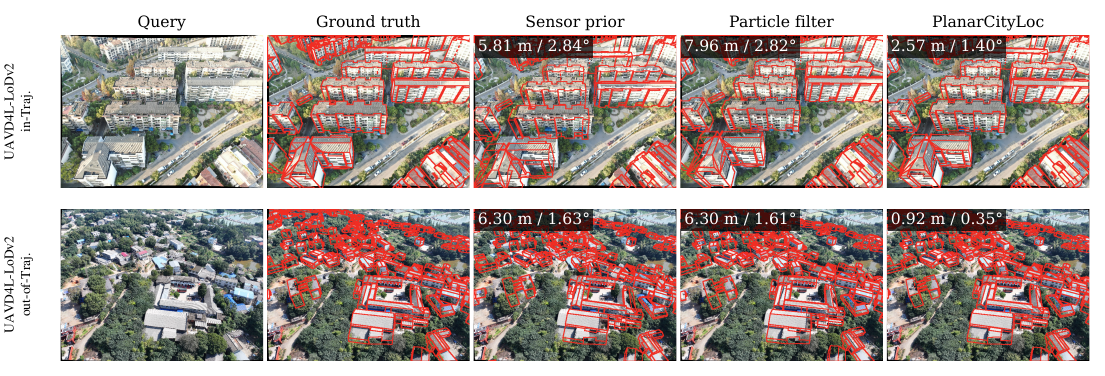}
\caption{\textbf{Pose estimates through PlanarCityLoc on UAVD4L-LoDv2.} Each row shows the query image and map overlays at the ground-truth, sensor-prior, particle-filter, and final selected poses. Labels give translation (m) / rotation ($^\circ$) errors.}
\label{fig:ambiguity}
\end{figure*}

\noindent \textbf{Qualitative analysis.}
\Cref{fig:iterations} shows that different methods can produce visually plausible alignments while yielding substantially different pose errors. Good image-space overlap alone is therefore insufficient to establish an accurate camera pose. The staged examples in \cref{fig:ambiguity} show how plane correspondences improve upon coarse alignment.

\subsection{Ablation Studies}
\label{sec:ablations}

We ablate components on fixed 100-query subsets per UAVD4L-LoDv2 split. Pose-selection and LoD comparisons use the full splits.

\noindent \textbf{Geometric abstraction (LoDs).}
\Cref{tab:ablation_lod} compares LoD1 and LoD3 on the complete UAVD4L-LoDv2 splits. Performance decreases with lower geometric detail. Coarser building shapes provide fewer faithful counterparts to image-side facade and roof planes, increasing the geometric gap and reducing correspondence accuracy.
\begin{table}[t]
\centering\small
\setlength{\tabcolsep}{2pt}
{\fontsize{7.272}{8.888}\selectfont
\begin{tabular*}{\columnwidth}{@{\extracolsep{\fill}}lccccc@{}}
\toprule
Map & 2m--2$^\circ$ & 3m--3$^\circ$ & 5m--5$^\circ$ & T.e.\,(m) & R.e.\,($^\circ$) \\
\midrule
LoD1 & \underline{23.8} / \underline{44.8} & \underline{54.1} / \underline{74.3} & \underline{82.2} / \underline{93.0} & \underline{2.85} / \underline{2.13} & \underline{1.26} / \underline{0.95} \\
LoD3 & \textbf{75.4} / \textbf{80.6} & \textbf{90.8} / \textbf{95.1} & \textbf{97.2} / \textbf{99.0} & \textbf{1.28} / \textbf{1.26} & \textbf{0.68} / \textbf{0.69} \\
\bottomrule
\end{tabular*}}
\caption{\textbf{LoD1 vs. LoD3.} Full UAVD4L-LoDv2 splits (in-Traj. / out-of-Traj.). Joint recalls (\%) and median errors.}
\label{tab:ablation_lod}
\end{table}

\begin{table}[t]
\centering\small
\setlength{\tabcolsep}{2pt}
\resizebox{\columnwidth}{!}{%
\begin{tabular}{lccccc}
\toprule
Selection rule & 2m--2$^\circ$ & 3m--3$^\circ$ & 5m--5$^\circ$ & T.e.\,(m) & R.e.\,($^\circ$) \\
\midrule
Particle filter only & 36.0 / 60.8 & 67.7 / 90.6 & 82.8 / 98.5 & 2.36 / 1.81 & 1.30 / 0.98 \\
Max PnP inliers & 65.8 / 74.8 & 84.1 / 91.1 & 95.4 / 98.2 & 1.49 / 1.37 & 0.82 / 0.76 \\
Max mask IoU & 67.4 / \underline{80.7} & 88.7 / \underline{95.3} & 96.8 / \textbf{99.1} & 1.54 / 1.32 & 0.79 / 0.75 \\
Max edge score & \textbf{76.6} / 79.6 & \underline{90.6} / 94.6 & \underline{97.1} / \underline{99.0} & \textbf{1.25} / 1.33 & \underline{0.69} / 0.77 \\
Max composite score & 72.3 / \textbf{81.2} & 89.7 / \textbf{96.0} & \textbf{97.2} / \underline{99.0} & 1.39 / \underline{1.30} & 0.72 / \underline{0.74} \\
Equal-weight ranks & \underline{75.4} / 80.6 & \textbf{90.8} / 95.1 & \textbf{97.2} / \underline{99.0} & \underline{1.28} / \textbf{1.26} & \textbf{0.68} / \textbf{0.69} \\
\bottomrule
\end{tabular}}
\caption{\textbf{Pose selection.} Full UAVD4L-LoDv2 splits (in-Traj. / out-of-Traj.), using identical candidate pools. Best/second-best values at shown precision are bold/underlined.}
\label{tab:ablation_pose}
\end{table}

\noindent \textbf{Refinement stages.}
The staged results in \cref{tab:main_results} and the examples in \cref{fig:ambiguity} show how coarse image-space alignment provides a useful initialization for explicit 3D refinement. Plane-based proposals substantially improve three splits, showing the complementary value of geometric correspondences beyond silhouette scoring.

\noindent \textbf{Pose selection.}
\looseness=1
\Cref{tab:ablation_pose} compares selection rules on identical candidate pools. Equal-weight rank aggregation gives the lowest median rotation error on both UAVD4L-LoDv2 splits and the lowest median translation error on out-of-Traj., while maintaining near-best strict recall. No single alignment cue consistently gives the best pose. These results support balancing complementary criteria through ranks, which avoid dependence on their different numerical scales without learning aggregation weights.

\begin{figure}[t]
\centering
\includegraphics[width=\linewidth]{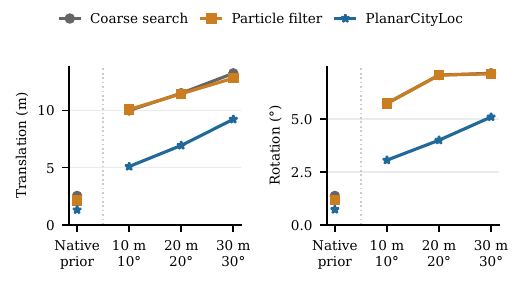}
\vspace{-1.75\baselineskip}
\caption{\textbf{Prior pose sensitivity on UAVD4L-LoDv2.} Curves show means of split-wise median errors for the same 300 queries per split. Synthetic priors use expanded search. Native-prior points are separate references.}
\vspace{-0.5\baselineskip}
\label{fig:noise_robustness}
\end{figure}

\noindent \textbf{Sensitivity to prior pose noise.}
We investigate sensitivity to the initial pose by increasing translation and orientation noise, with search ranges expanded to match the tested perturbations (\cref{fig:noise_robustness}). Larger perturbations lead to higher final errors. Nevertheless, plane refinement reduces both translation and rotation errors relative to particle filtering at every tested noise level, measured by the mean of the split-wise medians. The experiment tests refinement under substantial initial uncertainty beyond the released priors.

\noindent \textbf{Normal predictors.}
\Cref{tab:ablation_normals} compares geometry predictors on matched 100-query subsets. Each model supplies both depth and normals, while SAM~3 masks remain fixed. MoGe-2 outperforms the evaluated Metric3D v2 configurations in strict recall. Larger MoGe-2 backbones improve recall, especially on in-Traj. Main results use ViT-S.

\subsection{Limitations}
\label{sec:limitations}
PlanarCityLoc requires a useful coarse pose and sufficient reliable plane correspondences. Performance depends on the accuracy of the foundation-model predictions and the quality of the LoD geometry. In particular, approximating complex building shapes with LoD1 can make plane assignment noisier and reduce localization accuracy. The multi-stage pipeline combines neural predictions with repeated rendering and cannot yet operate onboard in real time without further optimization. Future work targets fast initialization and hardware-aware real-time optimization.
\begin{table}[t]
\centering\small
\setlength{\tabcolsep}{2pt}
\resizebox{\columnwidth}{!}{%
\begin{tabular}{llccccc}
\toprule
Model & Backbone & 2m--2$^\circ$ & 3m--3$^\circ$ & 5m--5$^\circ$ & T.e.\,(m) & R.e.\,($^\circ$) \\
\midrule
Metric3D v2 & ViT-S & 33.0 / 55.0 & 60.0 / 86.0 & 82.0 / 94.0 & 2.55 / 1.87 & 1.25 / 0.89 \\
Metric3D v2 & ViT-L & 31.0 / 58.0 & 62.0 / 78.0 & 87.0 / 90.0 & 2.41 / 1.80 & 1.33 / 0.96 \\
Metric3D v2 & ViT-G2 & 58.0 / 59.0 & 81.0 / 81.0 & 95.0 / 89.0 & 1.70 / 1.73 & 0.91 / 0.90 \\
MoGe-2 & ViT-S & 75.0 / \textbf{84.0} & 94.0 / \underline{94.0} & \textbf{98.0} / \underline{99.0} & 1.14 / 1.37 & 0.68 / 0.78 \\
MoGe-2 & ViT-B & \underline{82.0} / \underline{83.0} & \textbf{97.0} / 93.0 & \textbf{98.0} / 98.0 & \textbf{1.05} / \textbf{1.23} & \underline{0.65} / \underline{0.65} \\
MoGe-2 & ViT-L & \textbf{87.0} / \textbf{84.0} & \underline{96.0} / \textbf{97.0} & \underline{97.0} / \textbf{100.0} & \underline{1.09} / \underline{1.25} & \textbf{0.61} / \textbf{0.62} \\
\bottomrule
\end{tabular}}
\caption{\textbf{Geometry predictors.} Each model supplies both normals and depth. SAM3 masks are unchanged. All rows use the same 100 queries per UAVD4L-LoDv2 split (in-Traj. / out-of-Traj.). Recall uses joint thresholds. T.e./R.e. are medians. Bold/underline mark best/second-best at shown precision.}
\label{tab:ablation_normals}
\end{table}

\section{Conclusion}
\label{sec:conclusion}

We presented PlanarCityLoc, which grounds silhouette-based aerial localization in the 3D planar geometry of lightweight, textureless city models. By establishing plane correspondences and refining pose through PnP, our method achieves state-of-the-art localization recall among the evaluated off-the-shelf LoD-based methods without dataset-specific training. Our experiments demonstrate the value of planes for further improving pose estimation beyond image-space alignment. Effective refinement requires sufficient reliable plane correspondences. The current multi-stage pipeline is not yet real-time, motivating more efficient initialization and hardware-aware optimization for onboard deployment.

\section*{Acknowledgements}
Nikolas Rieger acknowledges support from the DAAD programme Konrad Zuse Schools of Excellence in Artificial Intelligence, funded by the Federal Ministry of Research, Technology and Space.
\FloatBarrier

\appendix
\renewcommand{\thefigure}{A\arabic{figure}}
\renewcommand{\thetable}{A\arabic{table}}
\renewcommand{\theHfigure}{supp.\arabic{figure}}
\renewcommand{\theHtable}{supp.\arabic{table}}
\setcounter{figure}{0}
\setcounter{table}{0}
\twocolumn[{\centering\Large\bfseries Supplementary Material\par\vspace{1em}}]
\newcommand{\mainresultstable}{\cref{tab:main_results}}
\setlength{\textfloatsep}{12pt plus 2pt minus 2pt}
\setlength{\dbltextfloatsep}{12pt plus 2pt minus 2pt}
\section{Overview}
\label{sec:supp_overview}

As demonstrated in the main paper, PlanarCityLoc {improves UAV localization through 3D planar constraints that address} the ambiguities of silhouette alignment.
Specifically, we first explain Computational Complexity in \cref{sec:supp_runtime} and discuss {Baseline Evaluation} in \cref{sec:supp_baseline_repro}.
We then present Extended Results in \cref{sec:supp_extended_quant}, along with Extended Ablation Studies in \cref{sec:supp_ablations} and a Robustness Analysis in \cref{sec:supp_robustness}.
Finally, we consider Limitations and Future Work in \cref{sec:supp_limitations} and Ethical Considerations in \cref{sec:supp_ethics}.

{Experiments use off-the-shelf SAM~3~\cite{carion2026sam} masks and MoGe-2 ViT-S~\cite{wang2025moge2} geometry unless stated otherwise. We report joint localization recall and median translation and rotation errors, following the main paper.}

\section{Computational Complexity}
\label{sec:supp_runtime}

{Our coarse stage follows the image-space search paradigm of prior work~\cite{zhu2025lod,peng2026lod}. PlanarCityLoc then introduces explicit 3D constraints through plane correspondences and PnP, avoiding another dense search over pose hypotheses.}

{On UAVD4L-LoDv2 in-Traj. and out-of-Traj., the rendering, matching, and PnP computations used in refinement take an average of 0.52 and 0.48\,s per query on an NVIDIA H100 NVL. On Swiss-EPFLv2 in-Place, these components take 0.23\,s on an NVIDIA A40. These component timings exclude one-time initialization, I/O, data transfers, plane extraction, and final pose selection.}

{Mean wall-clock latency is 6.42\,s per query on UAVD4L-LoDv2 in-Traj. (H100 NVL) and 8.75\,s on Swiss-EPFLv2 in-Place (A40), including image loading and transfers but excluding one-time initialization.}

{On the two UAVD4L-LoDv2 splits, refinement and final selection improve joint 2\,m--2$^\circ$ localization recall from 36.0\% to 75.4\% and from 60.8\% to 80.6\%, respectively (\cref{tab:supp_extended_quant}). This demonstrates the value of complementing coarse image-space alignment with correspondence-driven 3D pose updates.}

\section{{Baseline Evaluation}}
\label{sec:supp_baseline_repro}

{We distinguish published reference results from adapted evaluations in the main paper's \mainresultstable{}.}

{Our adapted evaluations share PlanarCityLoc's nvdiffrast~\cite{Laine2020diffrast} rasterizer and start from the released benchmark priors, with coarse pose search followed by particle filtering~\cite{zhu2025lod,peng2026lod}.}

{Ground-truth-rendered masks allow us to examine localization with ideal segmentation.}
{For LoD-Loc v2~\cite{zhu2025lod}, SAM~3~\cite{carion2026sam} additionally tests off-the-shelf segmentation. Published references retain their original configurations. Adapted evaluations of the public implementations use the same benchmark query splits with modified rendering and segmentation.}

\begingroup\space
\begin{table*}[t]
\centering\small
\setlength{\tabcolsep}{4pt}
{\fontsize{7.272}{8.888}\selectfont
\begin{tabular*}{\textwidth}{@{\extracolsep{\fill}}lllrrrrrrr@{}}
\toprule
Dataset / map & Split & Stage & Queries & 1m--1$^\circ$ & 2m--2$^\circ$ & 3m--3$^\circ$ & 5m--5$^\circ$ & T.e. & R.e. \\
\midrule
UAVD4L-LoDv2 / LoD3 & in-Traj. & Coarse search & 1604 & 1.4 & 27.6 & 59.9 & 79.4 & 2.62 & 1.49 \\
UAVD4L-LoDv2 / LoD3 & in-Traj. & Particle filter & 1604 & 3.1 & 36.0 & 67.7 & 82.8 & 2.36 & 1.30 \\
UAVD4L-LoDv2 / LoD3 & in-Traj. & PlanarCityLoc & 1604 & 36.3 & 75.4 & 90.8 & 97.2 & 1.28 & 0.68 \\
\midrule
UAVD4L-LoDv2 / LoD3 & out-of-Traj. & Coarse search & 2192 & 2.7 & 34.9 & 74.1 & 95.3 & 2.35 & 1.22 \\
UAVD4L-LoDv2 / LoD3 & out-of-Traj. & Particle filter & 2192 & 8.8 & 60.8 & 90.6 & 98.5 & 1.81 & 0.98 \\
UAVD4L-LoDv2 / LoD3 & out-of-Traj. & PlanarCityLoc & 2192 & 35.5 & 80.6 & 95.1 & 99.0 & 1.26 & 0.69 \\
\midrule
Swiss-EPFLv2 / LoD2 & in-Place & Coarse search & 712 & 2.5 & 15.4 & 30.2 & 48.7 & 5.04 & 1.72 \\
Swiss-EPFLv2 / LoD2 & in-Place & Particle filter & 712 & 5.2 & 24.9 & 41.6 & 61.8 & 3.54 & 1.54 \\
Swiss-EPFLv2 / LoD2 & in-Place & PlanarCityLoc & 712 & 4.4 & 20.6 & 39.0 & 59.7 & 3.70 & 1.90 \\
\midrule
Swiss-EPFLv2 / LoD2 & out-of-Place & Coarse search & 379 & 1.1 & 11.3 & 23.2 & 57.5 & 4.42 & 1.41 \\
Swiss-EPFLv2 / LoD2 & out-of-Place & Particle filter & 379 & 1.3 & 14.5 & 31.1 & 70.2 & 3.99 & 1.23 \\
Swiss-EPFLv2 / LoD2 & out-of-Place & PlanarCityLoc & 379 & 7.9 & 48.5 & 72.3 & 89.4 & 1.98 & 1.21 \\
\midrule
UAVD4L-LoDv2 / LoD1 & in-Traj. & Coarse search & 1604 & 2.1 & 19.3 & 45.3 & 72.5 & 3.24 & 1.60 \\
UAVD4L-LoDv2 / LoD1 & in-Traj. & Particle filter & 1604 & 2.7 & 22.0 & 51.1 & 73.8 & 2.94 & 1.38 \\
UAVD4L-LoDv2 / LoD1 & in-Traj. & PlanarCityLoc & 1604 & 3.6 & 23.8 & 54.1 & 82.2 & 2.85 & 1.26 \\
\midrule
UAVD4L-LoDv2 / LoD1 & out-of-Traj. & Coarse search & 2192 & 2.3 & 24.3 & 56.9 & 86.4 & 2.75 & 1.25 \\
UAVD4L-LoDv2 / LoD1 & out-of-Traj. & Particle filter & 2192 & 3.9 & 40.8 & 75.2 & 92.8 & 2.23 & 1.03 \\
UAVD4L-LoDv2 / LoD1 & out-of-Traj. & PlanarCityLoc & 2192 & 8.3 & 44.8 & 74.3 & 93.0 & 2.13 & 0.95 \\
\bottomrule\end{tabular*}}
\caption{Full-split staged results. Recalls are joint percentages. Translation/rotation errors are medians in meters/degrees. Every stage is evaluated on the same queries within a split.}
\label{tab:supp_extended_quant}
\end{table*}
\endgroup

\section{Extended Results}
\label{sec:supp_extended_quant}

\begin{figure*}[t]
\centering
\includegraphics[width=\textwidth]{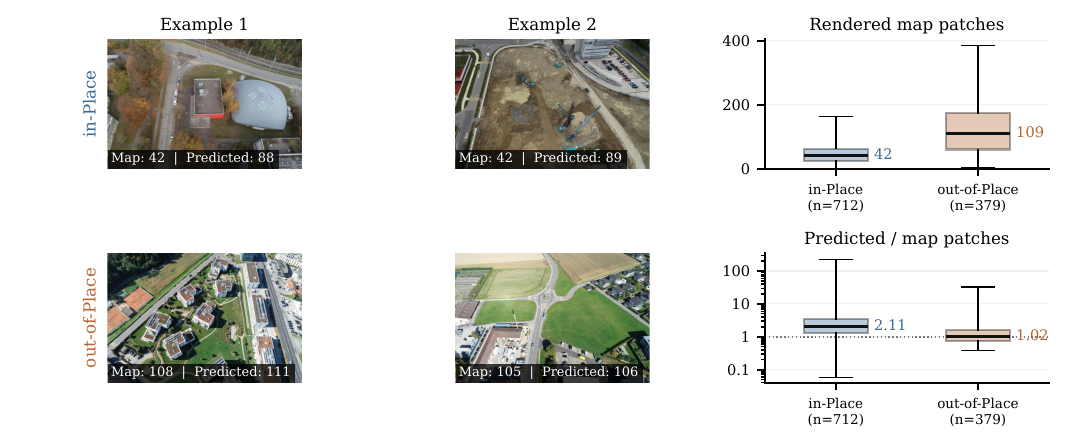}
\caption{{\textbf{Scene context and planar support on Swiss-EPFLv2.} Left: views near each split's median map-patch count and predicted-to-rendered ratio. Right: full-split distributions at the particle-filter pose. Boxes show interquartile ranges, lines show medians, and whiskers show the full range. The ratio axis is logarithmic.}}
\label{fig:supp_swiss_scenes}
\end{figure*}

\noindent\textbf{Per-split breakdowns.}
We report the complete staged evaluation in \cref{tab:supp_extended_quant}.
This evaluation details the performance of the cost volume, particle filter, and our plane matching approach across all benchmark splits.
{The full-split results show substantial gains from plane matching on both UAVD4L-LoDv2 splits and Swiss-EPFLv2 out-of-Place, supporting the use of 3D constraints beyond silhouette agreement.}
We also report the full-split UAVD4L-LoDv2 LoD1 ablation and both Swiss-EPFLv2 splits with LoD2 maps.
{Swiss-EPFLv2 in-Place is more challenging for refinement (\cref{tab:supp_extended_quant}). \Cref{fig:supp_swiss_scenes} compares scene examples and planar support. At the particle-filter pose, this split has fewer rendered map patches and a larger predicted-to-rendered patch-count ratio than out-of-Place. The examples contrast sparse building coverage and construction areas with multi-building scenes. Lower map support and greater patch-count imbalance suggest more challenging image-to-map plane association.}

\begin{figure*}[t]
\centering
\includegraphics[width=0.92\textwidth]{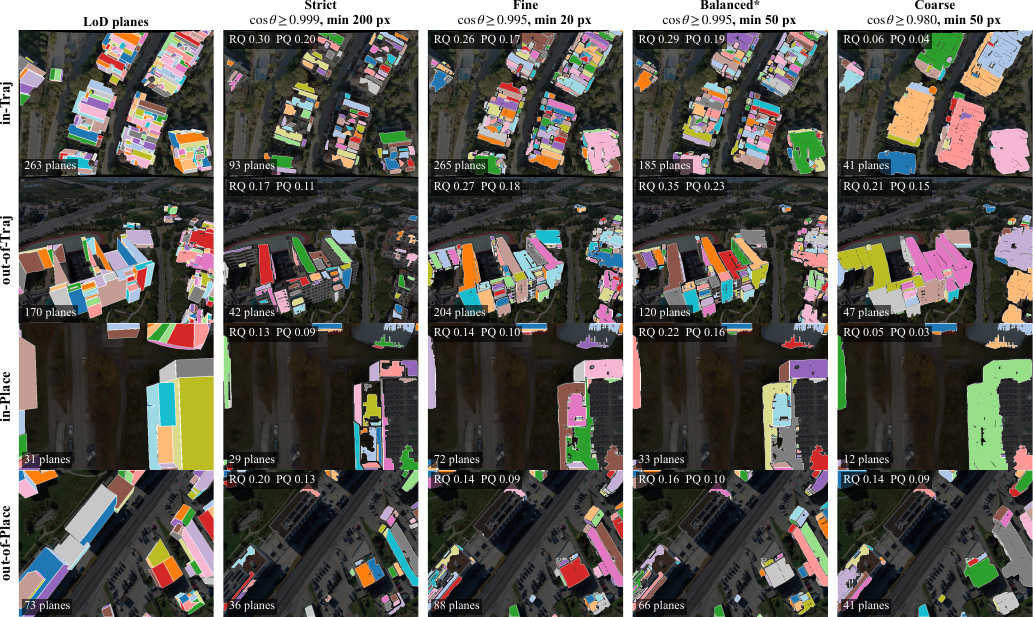}
\caption{\textbf{Plane extraction qualitative comparison.} LoD primitives against predicted plane extractions. {Patch granularity varies with normal consistency and minimum region size.} RQ and PQ denote Recognition Quality and Panoptic Quality, respectively.}
\label{fig:planes_lod_vs_pred_supp}
\vspace{-0.5\baselineskip}
\end{figure*}

\noindent\textbf{Plane extraction against LoD primitives.}
{\Cref{fig:planes_lod_vs_pred_supp}} compares LoD planes rendered from the mesh with planes segmented from predicted normals.
LoD planes follow the mesh tessellation and remain largely insensitive to the normal-consistency threshold.
In contrast, predicted planes depend {on normal and depth consistency as well as minimum region size}.
{Fine settings can fragment a facade, while coarse settings can merge distinct surfaces. The examples illustrate this trade-off in the resulting planar representation.}

\begin{figure*}[t]
\centering
\includegraphics[width=0.88\textwidth]{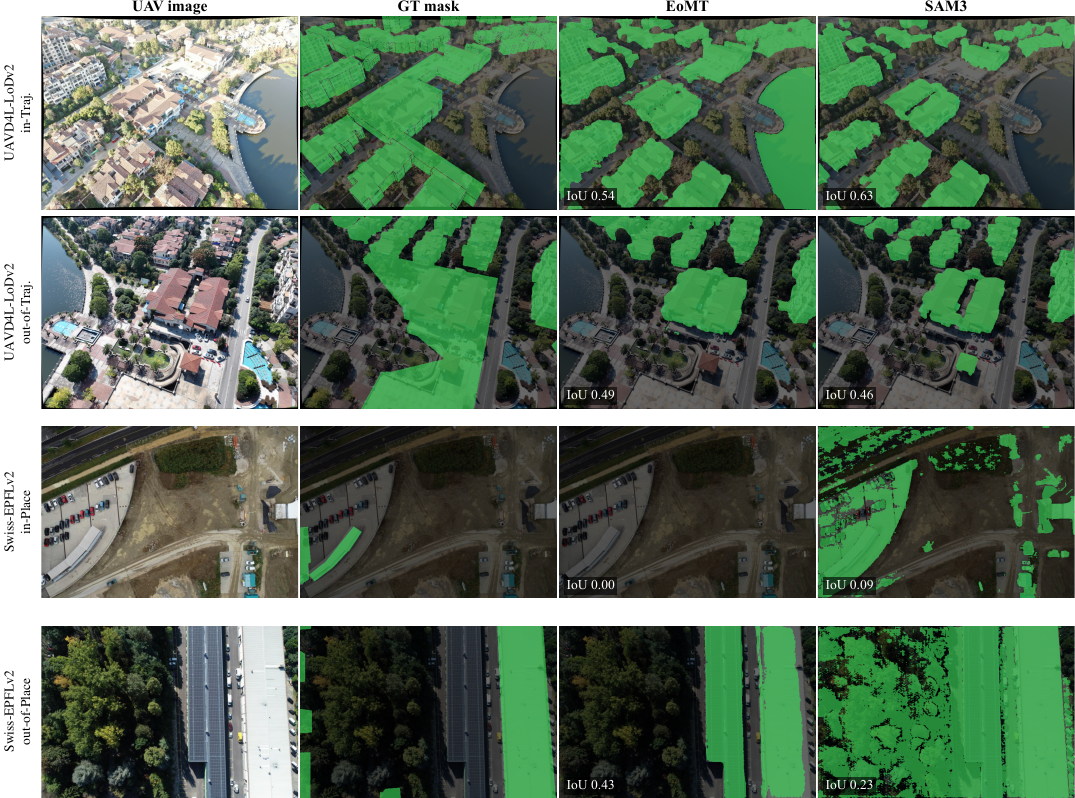}
\caption{\textbf{Mask segmentation {examples}.} {Query images and predicted masks illustrate incomplete building coverage and non-building regions included by off-the-shelf segmentation. Such errors can weaken coarse alignment and the planar evidence available for refinement.}}
\label{fig:mask_segmentation_supp_bad}
\end{figure*}

\noindent\textbf{Failure cases.}
{Planar refinement requires useful image-to-map alignment. \Cref{fig:mask_segmentation_supp_bad} illustrates initialization failures. Off-the-shelf segmentation under domain shift can miss small buildings, misclassify lakes or parking lots, or produce noisy masks, favoring incorrect cost-volume hypotheses.}
{Disjoint query and rendered masks then supply no overlap-based distinction between nearby hypotheses, weakening the initialization for plane matching.}
Beyond pure segmentation errors, matching can also degrade due to weak building visibility, ambiguous facade normals from repetitive architecture, or temporal mismatches between outdated LoD geometry (e.g., ground artifacts modeled as buildings) and the true UAV capture.
{When geometric and mask cues disagree strongly, reliable confidence estimation is needed to avoid incorrect pose updates.}

\section{Extended Ablation Studies}
\label{sec:supp_ablations}

{We ablate core design choices on the same fixed 100-query subset per UAVD4L-LoDv2 split, holding other pipeline settings constant.}

\noindent\textbf{Coarse scoring components.}
{\Cref{tab:supp_ablation_score}} demonstrates the impact of different geometric cues {during coarse search and particle filtering}.
Relying solely on mask intersection-over-union (IoU) provides a weak signal under occlusion.
{Edges add boundary constraints, while depth helps distinguish hypotheses with similar silhouettes but different spatial configurations. Combining all three cues gives the highest strict recall on both subsets, supporting their complementary role in initialization.}
\begingroup\space
\begin{table}[t]
\centering\small
\setlength{\tabcolsep}{2pt}
{\fontsize{7.272}{8.888}\selectfont
\begin{tabular*}{\columnwidth}{@{\extracolsep{\fill}}lccccc@{}}
\toprule
Configuration & 2m--2$^\circ$ & 3m--3$^\circ$ & 5m--5$^\circ$ & T.e.\,(m) & R.e.\,($^\circ$) \\
\midrule
Mask only & 62.0 / 66.0 & 87.0 / 82.0 & \underline{95.0} / 94.0 & 1.39 / 1.50 & 0.81 / 0.81 \\
Mask + edges & \underline{74.0} / \underline{76.0} & \underline{92.0} / \underline{91.0} & \textbf{98.0} / \underline{97.0} & \underline{1.20} / \underline{1.42} & \underline{0.75} / \textbf{0.71} \\
Mask + depth & 60.0 / 71.0 & 80.0 / 89.0 & 92.0 / \underline{97.0} & 1.66 / 1.49 & 0.95 / \underline{0.74} \\
All three cues & \textbf{75.0} / \textbf{84.0} & \textbf{94.0} / \textbf{94.0} & \textbf{98.0} / \textbf{99.0} & \textbf{1.14} / \textbf{1.37} & \textbf{0.68} / 0.78 \\
\bottomrule
\end{tabular*}}
\caption{\textbf{Coarse scoring components.} Mask, edge, and depth cues in coarse search and particle filtering. Other stages are unchanged. Matched 100-query UAVD4L-LoDv2 subsets (in-Traj. / out-of-Traj.). Joint recalls (\%) and median errors. Best/second-best values are bold/underlined.}
\label{tab:supp_ablation_score}
\end{table}
\endgroup

\noindent\textbf{Plane extraction criteria.}
{\Cref{tab:supp_ablation_extraction}} explores the logic for grouping predicted surface normals into planar primitives.
Grouping by normal consistency alone captures facades but {can merge adjacent surfaces with similar orientations}.
{Depth continuity helps separate these surfaces, but alone does not distinguish their orientation. Combining both criteria gives the highest strict recall and lowest median translation error on both subsets, while normal consistency alone remains competitive at looser thresholds.}
\begingroup\space
\begin{table}[t]
\centering\small
\setlength{\tabcolsep}{2pt}
{\fontsize{7.272}{8.888}\selectfont
\begin{tabular*}{\columnwidth}{@{\extracolsep{\fill}}lccccc@{}}
\toprule
Configuration & 2m--2$^\circ$ & 3m--3$^\circ$ & 5m--5$^\circ$ & T.e.\,(m) & R.e.\,($^\circ$) \\
\midrule
Normal consistency & \underline{74.0} / \underline{79.0} & \textbf{94.0} / \textbf{96.0} & \textbf{98.0} / \textbf{100.0} & \underline{1.30} / \underline{1.41} & \underline{0.78} / 0.79 \\
Depth continuity & 31.0 / 64.0 & \underline{72.0} / 92.0 & \underline{93.0} / 97.0 & 2.36 / 1.67 & 1.25 / \textbf{0.77} \\
Normals + depth & \textbf{75.0} / \textbf{84.0} & \textbf{94.0} / \underline{94.0} & \textbf{98.0} / \underline{99.0} & \textbf{1.14} / \textbf{1.37} & \textbf{0.68} / \underline{0.78} \\
\bottomrule
\end{tabular*}}
\caption{\textbf{Plane extraction criteria.} Normal and depth consistency for predicted and rendered planes. Matched 100-query UAVD4L-LoDv2 subsets (in-Traj. / out-of-Traj.). Joint recalls (\%) and median errors. Best/second-best values are bold/underlined.}
\label{tab:supp_ablation_extraction}
\end{table}
\endgroup

\noindent\textbf{PnP matching strategy.}
{\Cref{tab:supp_ablation_matching}} evaluates our sequential alignment criteria during PnP optimization.
{Overlap voting accommodates fragmented image patches, while centroid consistency establishes more localized correspondences. Applying both matchers sequentially combines these complementary cues and gives the highest strict recall on both subsets. Overlap-only matching remains stronger at some looser out-of-Traj. thresholds.}
\begingroup\space
\begin{table}[t]
\centering\small
\setlength{\tabcolsep}{2pt}
{\fontsize{7.272}{8.888}\selectfont
\begin{tabular*}{\columnwidth}{@{\extracolsep{\fill}}lccccc@{}}
\toprule
Configuration & 2m--2$^\circ$ & 3m--3$^\circ$ & 5m--5$^\circ$ & T.e.\,(m) & R.e.\,($^\circ$) \\
\midrule
Overlap only & 54.0 / \underline{76.0} & \underline{89.0} / \textbf{96.0} & 96.0 / \underline{98.0} & 1.80 / \underline{1.41} & 1.02 / 0.86 \\
Centroid only & \underline{64.0} / 72.0 & 85.0 / 91.0 & \underline{97.0} / \textbf{99.0} & \underline{1.50} / 1.59 & \underline{0.91} / \underline{0.79} \\
Overlap + centroid & \textbf{75.0} / \textbf{84.0} & \textbf{94.0} / \underline{94.0} & \textbf{98.0} / \textbf{99.0} & \textbf{1.14} / \textbf{1.37} & \textbf{0.68} / \textbf{0.78} \\
\bottomrule
\end{tabular*}}
\caption{\textbf{PnP matching strategies.} Overlap, centroid, and sequential correspondences. Matched 100-query UAVD4L-LoDv2 subsets (in-Traj. / out-of-Traj.). Joint recalls (\%) and median errors. Best/second-best values are bold/underlined.}
\label{tab:supp_ablation_matching}
\end{table}
\endgroup

\begin{figure*}[!t]
\centering
\includegraphics[width=0.8\textwidth]{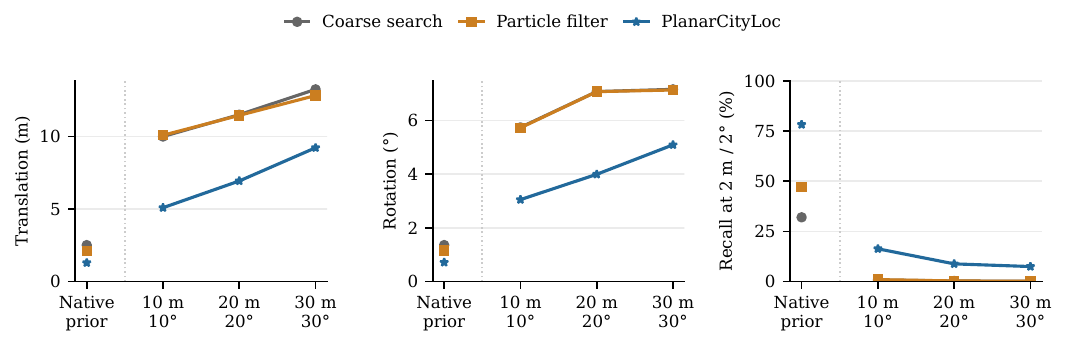}
\caption{{\textbf{Comprehensive prior noise evaluation.} Joint recall and median errors under synthetic pose perturbations on UAVD4L-LoDv2, using 300 queries per split and expanded search ranges. Native-prior points are separate references on the same queries.}}
\label{fig:supp_noise_all}
\end{figure*}

\begin{figure*}[!t]
\centering
\includegraphics[width=\textwidth]{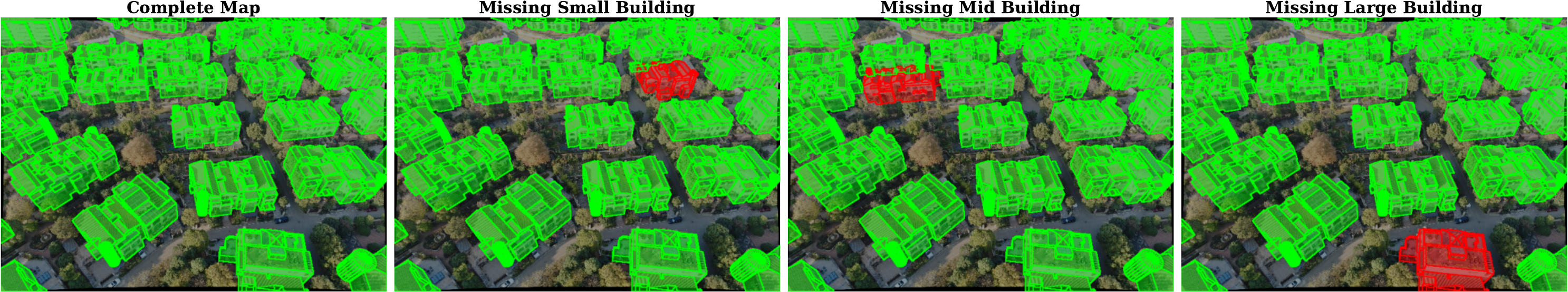}
\caption{\textbf{Visual examples of map completeness.} Overlays of rendered LoD {geometry for the intact map and representative removal conditions. Groups are selected by projected area, not physical building volume.}}
\label{fig:supp_map_completeness_visual}
\end{figure*}

\noindent\textbf{Outlier rejection.}
{PnP-RANSAC~\cite{fischler1981random} rejects inconsistent correspondences and refines the pose on the inliers. Updating correspondences as the pose changes allows subsequent iterations to use the improved image-to-map alignment.}

\noindent\textbf{Pose selection.}
Reprojection and centroid errors are obtained from PnP correspondences and are not computed for the particle-filter candidate. For each criterion, candidates without a measurement rank below those with valid measurements. Equal criterion values receive average ranks. Equal aggregate ranks retain the earlier candidate, with the particle-filter estimate considered first. These conventions also apply to the full-split pose-selection comparison in the main paper.

\noindent\textbf{Building segmenters.}
{\Cref{tab:supp_ablation_segmenter}} compares the impact of the upstream vision model, with visual examples shown in \cref{fig:mask_segmentation_supp_bad}.
While the geometry prediction (MoGe-2) remains fixed, swapping the SAM3~\cite{carion2026sam} mask module for EoMT~\cite{kerssies2025eomt} severely degrades performance.
{Reliable building masks are therefore important for both coarse alignment and subsequent plane matching.}
\begingroup\space
\begin{table}[t]
\centering\small
\setlength{\tabcolsep}{2pt}
{\fontsize{7.272}{8.888}\selectfont
\begin{tabular*}{\columnwidth}{@{\extracolsep{\fill}}lccccc@{}}
\toprule
Configuration & 2m--2$^\circ$ & 3m--3$^\circ$ & 5m--5$^\circ$ & T.e.\,(m) & R.e.\,($^\circ$) \\
\midrule
SAM3 & \textbf{75.0} / \textbf{84.0} & \textbf{94.0} / \textbf{94.0} & \textbf{98.0} / \textbf{99.0} & \textbf{1.14} / \textbf{1.37} & \textbf{0.68} / \textbf{0.78} \\
EoMT & \underline{40.0} / \underline{48.0} & \underline{54.0} / \underline{61.0} & \underline{69.0} / \underline{70.0} & \underline{2.51} / \underline{2.26} & \underline{1.46} / \underline{1.06} \\
\bottomrule
\end{tabular*}}
\caption{\textbf{Building segmenters.} SAM~3 and EoMT with fixed MoGe-2 ViT-S geometry. Matched 100-query UAVD4L-LoDv2 subsets (in-Traj. / out-of-Traj.). Joint recalls (\%) and median errors. Best/second-best values are bold/underlined.}
\label{tab:supp_ablation_segmenter}
\end{table}
\endgroup

\section{Robustness Analysis}
\label{sec:supp_robustness}

\noindent\textbf{Sensitivity to prior pose noise.}
We summarize median pose error against injected prior noise in \cref{fig:supp_noise_all}.
{We evaluate synthetic priors on 300 fixed queries per UAVD4L-LoDv2 split. Translation and angular components are uniformly perturbed within bounds of 10, 20, or 30 meters/degrees, with pitch and roll limited to $\pm5^\circ$. Search ranges are expanded to accommodate the increased uncertainty. Native-prior results are shown separately.}

{Larger initial errors reduce localization accuracy. Plane-based refinement nevertheless lowers median translation and rotation errors relative to particle filtering at every tested perturbation level. At the largest offsets, absolute recall remains low, highlighting the importance of a useful initialization even when 3D constraints improve the estimate.}

\noindent\textbf{Sensitivity to map completeness.}
\begingroup\space
\begin{table}[t]
\centering\small
\setlength{\tabcolsep}{2pt}
{\fontsize{7.272}{8.888}\selectfont
\begin{tabular*}{\columnwidth}{@{\extracolsep{\fill}}lccccc@{}}
\toprule
Configuration & 2m--2$^\circ$ & 3m--3$^\circ$ & 5m--5$^\circ$ & T.e.\,(m) & R.e.\,($^\circ$) \\
\midrule
Intact map & \underline{75.0} / \textbf{84.0} & \textbf{94.0} / 94.0 & \textbf{98.0} / \textbf{99.0} & \textbf{1.14} / \textbf{1.37} & \textbf{0.68} / 0.78 \\
Remove smallest & \textbf{79.0} / \underline{82.0} & \textbf{94.0} / \textbf{98.0} & \textbf{98.0} / \textbf{99.0} & \underline{1.18} / \underline{1.41} & \textbf{0.68} / \textbf{0.73} \\
Remove middle & 71.0 / 77.0 & \underline{89.0} / \underline{95.0} & \underline{94.0} / \textbf{99.0} & 1.41 / 1.46 & \underline{0.72} / 0.81 \\
Remove largest & 52.0 / 69.0 & 70.0 / 84.0 & 84.0 / \underline{95.0} & 1.87 / 1.47 & 1.02 / \underline{0.76} \\
\bottomrule
\end{tabular*}}
\caption{\textbf{Map completeness.} Matched 100-query UAVD4L-LoDv2 subsets (in-Traj. / out-of-Traj.). Groups of nearby mesh surfaces are removed by projected area. Queries without sufficient groups retain the intact map. Joint recalls (\%) and median errors. Best/second-best values are bold/underlined.}
\label{tab:supp_map_ablation}
\end{table}
\endgroup
We investigate the sensitivity of our pipeline to {missing structures} in the LoD model in {\cref{tab:supp_map_ablation}}.
For each frame, we selectively remove {a visible mesh-derived group of nearby surfaces} before rendering the map prior.
{Removal size is defined by projected area at the initial pose, with each group approximating a visible structure rather than a labeled building instance.}
{Removing small or medium groups has mixed effects, while removing the largest group lowers strict recall on both splits. This supports the importance of large visible structures as geometric anchors for localization.}
{\Cref{fig:supp_map_completeness_visual}} visualizes these scenarios.

\section{Limitations and Future Work}
\label{sec:supp_limitations}

{PlanarCityLoc depends on reliable coarse initialization. Poor masks or inaccurate GPS priors can lead to incorrect alignment that planar refinement cannot recover (\cref{fig:mask_segmentation_supp_bad}). Localization also degrades with incomplete or poorly modeled LoD geometry (\cref{tab:supp_map_ablation}). Better upstream predictions and robust initialization could mitigate these limitations. Temporal filtering could disambiguate repetitive structures through motion cues. Efficient initialization and hardware-aware optimization remain necessary for onboard real-time operation.}

\section{Ethical Considerations}
\label{sec:supp_ethics}

\noindent\textbf{Geographic bias and reliability.}
Our zero-shot evaluation is geographically constrained to specific urban environments, and performance may vary across different architectural styles.
Lacking hard accuracy guarantees, the system requires rigorous fallbacks before deployment in safety-critical autonomous flights.

\noindent\textbf{Privacy and data usage.}
{Public datasets and textureless maps do not eliminate privacy risks. Query images and pretrained models may contain sensitive information, and downstream use must respect applicable data and model terms. Swiss-EPFL map data are provided by the Federal Office of Topography swisstopo.}

\noindent\textbf{Dual-use potential.}
{Although intended for civilian inspection, aerial localization can enable surveillance and requires safeguards against misuse.}

\end{document}